\documentclass[conference]{IEEEtran}
\IEEEoverridecommandlockouts

\usepackage{times}
\usepackage{latexsym}
\usepackage{cite}
\usepackage{url}
\usepackage{algorithm}
\usepackage{algpseudocode}
\usepackage[T1]{fontenc}

\usepackage[utf8]{inputenc}

\usepackage{microtype}
\usepackage{inconsolata}

\usepackage{graphicx}
\usepackage{amsmath}
\usepackage{amssymb}
\usepackage{booktabs}
\usepackage{multirow}
\usepackage{siunitx}
\usepackage{tabularx}
\usepackage{makecell}
\usepackage{array}
\usepackage[caption=false,font=footnotesize]{subfig}

\usepackage{xcolor}
\definecolor{navyblue}{RGB}{0,0,128}

\newcommand{\ms}[2]{\ensuremath{#1_{\scriptstyle\pm #2}}}
\newcommand{\second}[1]{\underline{\textit{#1}}}
\newcommand{\lord}[2]{{\bfseries\boldmath\ms{#1}{#2}}}

\title{Too Sure to Be Safe: Model Calibration for Reliable Log Anomaly Detection
\thanks{Accepted at the 2026 IEEE International Conference on Data Mining.}
}

\author{
\IEEEauthorblockN{
Bin Li$^{1*}$,
Dongdong Wang$^{2*}$,
Siyang Lu$^{3\dagger}$
}

\IEEEauthorblockA{
$^{1}$School of Cyberspace Science and Technology, Beijing Jiaotong University, Beijing, China\\
$^{2}$University of Florida, Gainesville, FL, USA\\
$^{3}$School of Computer Science and Technology, Beijing Jiaotong University, Beijing, China\\
24120434@bjtu.edu.cn, dongdongwang@ufl.edu, sylu@bjtu.edu.cn
}

\thanks{$^{*}$These authors contributed equally to this work.}
\thanks{$^{\dagger}$Corresponding author: Siyang Lu (sylu@bjtu.edu.cn).}
}

\begin{document}
\maketitle
\begin{abstract}


Online log anomaly detection is critical for maintaining the reliability of large-scale computing systems. Although recent language model-based log anomaly detectors achieve strong detection performance, their confidence estimates remain poorly calibrated. We show that these detectors frequently assign excessive confidence to incorrect predictions, particularly for anomalous logs under severe class imbalance. Moreover, confidence on erroneous predictions remains persistently high even when conventional calibration metrics indicate good calibration, creating a critical reliability gap for operational monitoring systems. To address this issue, we propose Log Reconstruction and Distance (LoRD), a lightweight post-hoc calibration framework for reliable log anomaly detection. LoRD learns prediction-route-specific reliability models from latent representations of correctly classified validation samples and estimates prediction reliability through route-wise reconstruction distances. Based on the estimated reliability, LoRD selectively recalibrates high-risk predictions to suppress overconfident errors while preserving reliable predictions. Extensive experiments on four large-scale log benchmark datasets and multiple language model-based detectors demonstrate that LoRD consistently improves confidence reliability and substantially reduces overconfident anomaly-related errors without sacrificing anomaly detection performance.


\end{abstract}

\section{Introduction}



Large-scale data systems are increasingly critical in the era of artificial intelligence and high-performance computing. Maintaining system reliability relies heavily on continuous log monitoring, making log anomaly detection essential for identifying abnormal events and preventing system failures. Existing log anomaly detection approaches have achieved substantial progress in semantic log analysis and anomalous event detection~\cite{han2022skdlog, lu2023ssdlog, lu2023black, nedelkoski2020self}.

Current methods for log anomaly detection can generally be categorized into two paradigms~\cite{le2022log}. The first models normal log patterns and identifies deviations from expected behavior as anomalies. The other formulates log anomaly detection as a supervised classification problem, where models directly predict whether a log sequence is normal or anomalous. While the former typically requires less labeled data, the latter often achieves stronger benchmark~\cite{landauer2024critical, zhu2023loghub} performance but relies heavily on high-quality annotations and reliable preprocessing pipelines. In this paper, we focus on supervised classification-based log anomaly detectors and investigate the reliability of their confidence estimation.

However, strong classification performance does not guarantee reliable decision-making. In log anomaly detection, prediction errors are asymmetric in their operational impact~\cite{wu2023towards}. A normal log sequence that is incorrectly classified as anomalous may trigger a false alarm, causing additional inspection costs and potentially contributing to alert fatigue. While such errors are undesirable, they still expose suspicious cases to operators or downstream monitoring modules. By contrast, an anomalous log sequence that is incorrectly classified as normal constitutes a missed anomaly, allowing the underlying abnormal behavior to bypass the detection pipeline. This risk becomes more severe when the detector assigns high confidence to the incorrect normal prediction, as the system may treat the decision as reliable and suppress further intervention. Therefore, high-confidence false negatives constitute a central reliability risk for supervised log anomaly detectors.

We first examine the confidence behavior of representative supervised log anomaly detectors. The results reveal a recurring reliability issue: detectors can be confidently wrong. Both false positives and false negatives often receive confidence scores comparable to correctly classified samples, suggesting that detector confidence is not consistently aligned with prediction correctness. This issue is not simply resolved by increasing model complexity. More complex detectors may obtain higher accuracy and more favorable aggregate calibration scores, yet their confidence distributions often remain highly concentrated near one, leaving incorrect predictions overconfident. We also find that anomalous samples tend to exhibit larger calibration errors than normal samples. These observations indicate that strong benchmark performance can still conceal important reliability risks, particularly when anomalous log sequences are confidently predicted as normal.

To mitigate this risk, we design a selector-based calibration framework for supervised log anomaly detectors. Given the output of a pre-trained log anomaly detector, the selector estimates prediction reliability from the detector’s hidden representation. Specifically, for each input, we use the detector's  hidden representation as the basis for reliability estimation. Since predictions assigned to the normal class and those assigned to the anomalous class exhibit different error patterns, we construct two prediction routes according to the detector's output label. For each route, an autoencoder is trained on the hidden representations of correctly classified validation samples, and its reconstruction error is used as a reliability score.
Although the two routes follow the same autoencoder-based scoring principle, their calibration objectives are asymmetric. The predicted-normal route prioritizes potential false negatives, while the predicted-anomalous route is treated more conservatively to avoid weakening reliable anomaly predictions. This design reflects the asymmetric risk of log anomaly detection, where missed anomalies are typically more critical than false alarms.

\begin{figure*}[t]
    \centering
    \includegraphics[width=0.99\textwidth]{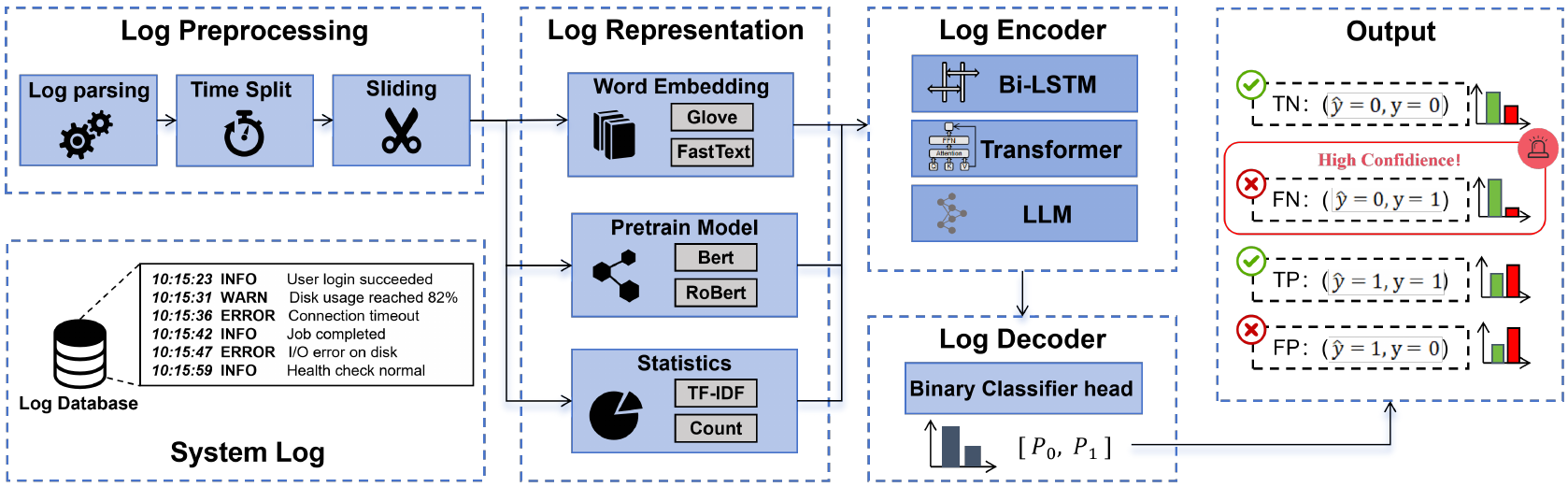}
    \caption{Overview of existing log anomaly detection frameworks. Logs are preprocessed, embedded into semantic representations, encoded using sequential models such as Bi-LSTM or Transformer, and finally used for anomaly prediction.}
    \label{fig:workflow}
\end{figure*}

In this work, we first show that supervised log anomaly detectors can remain overconfident on misclassified samples despite strong benchmark performance. 
We then propose a selector-based calibration framework that estimates prediction reliability from the detector’s predictions and hidden representations.
Finally, we demonstrate across multiple datasets and representative detectors that our method achieves a task-oriented calibration trade-off, yielding more reliable confidence estimates for anomalous samples, especially those confidently predicted as normal, while keeping the calibration degradation on normal samples within an acceptable range.

Our contributions are summarized as follows:

\begin{itemize}

\item To the best of our knowledge, this is the first work to systematically investigate reliability and confidence calibration in language model-based log anomaly detection. We reveal a persistent overconfidence phenomenon in which state-of-the-art detectors assign excessively high confidence to erroneous anomaly predictions despite exhibiting strong performance under conventional calibration metrics. Through extensive analyses across history window length, LoRA rank, model complexity, and class-imbalance mitigation strategies, we show that Confidence on Error (CoE) remains consistently high across models and settings, exposing a critical reliability gap that is overlooked by existing calibration metrics.

\item We propose \textbf{LoRD} (Log Reconstruction and Distance), a lightweight post-hoc calibration framework for log anomaly detection. LoRD learns route-specific reconstruction models that characterize the reliable conditional latent distributions of normal and anomalous predictions, and leverages reconstruction distances as reliability signals for tri-region confidence calibration. By selectively preserving reliable predictions while suppressing overconfident errors, LoRD directly targets the reliability challenge identified in modern log anomaly detectors.

\item We conduct extensive experiments on four large-scale log benchmark datasets using diverse language model-based log anomaly detectors with varying architectures and model scales. Experimental results demonstrate that LoRD consistently improves reliability and calibration quality across datasets and detectors, effectively reducing overconfident mispredictions while preserving detection performance. These findings highlight the robustness, effectiveness, and generalizability of LoRD for reliability-aware log anomaly detection.
\end{itemize}

\section{Related Work}


\subsection{Log Anomaly Detection}


Researchers have extensively applied sequence modeling techniques to system log analysis, where log anomaly detection has emerged as a critical task. Driven by advances in deep neural networks, increasingly powerful models have substantially improved the accuracy and robustness of log anomaly detection.

Log anomaly detection aims to identify abnormal system behaviors from log data. Existing studies mainly follow two lines~\cite{le2022log}: i) semi-supervised methods and ii) supervised methods.
Semi-supervised methods usually learn normal log patterns from historical logs and identify samples that deviate from the learned normality as anomalies. 
DeepLog~\cite{du2017deeplog} is a representative method in this line, which models normal log sequences with an LSTM and detects anomalies based on deviations from predicted execution patterns. 
LogAnomaly~\cite{meng2019loganomaly} further models sequential and quantitative anomalies by incorporating semantic log representations.
PLELog~\cite{yang2021plelog} reduces reliance on manual labeling through probabilistic label estimation and semantic representation learning for log anomaly detection.

\begin{figure*}[t]
\centering
\includegraphics[width=0.99\textwidth]{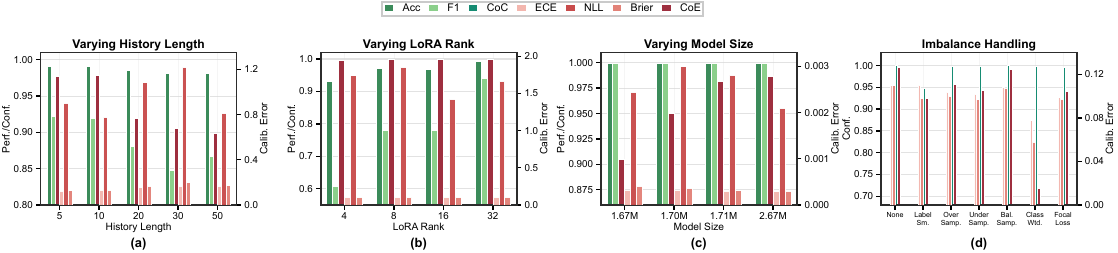}
\caption{Ablation analysis of factors affecting confidence calibration and error confidence in log anomaly detection. Panels (a)--(c) report Acc, F1, and CoE on the left axis and ECE, NLL, and Brier score on the right axis. Panel (d) reports abnormal-class ECE, Brier score, CoC, and CoE for imbalance handling strategies. Green bars denote higher-is-better metrics, while red bars denote lower-is-better metrics. Label Sm., Over Samp., Under Samp., Bal. Samp., and Class Wtd. denote Label Smoothing, Over-Sampler, Under-Sampler, Balanced Sampler, and Class-Weighted Loss, respectively.}
\label{fig:ablation_studies}
\end{figure*}


\begin{figure*}[t]
\centering
\includegraphics[width=0.99\textwidth]{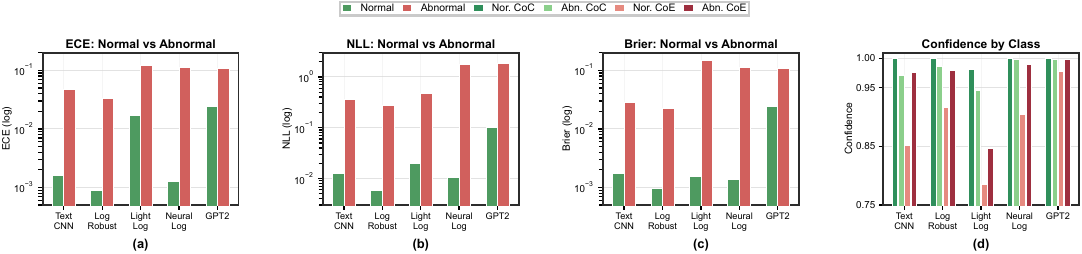}
\caption{Class-wise calibration comparison on the BGL dataset. Panels (a)--(c) compare normal-class and abnormal-class ECE, NLL, and Brier score using a logarithmic axis. Panel (d) compares class-wise confidence on correct predictions and erroneous predictions.}
\label{fig:supp_bgl_classwise}
\end{figure*}


Supervised methods treat log anomaly detection as a binary classification task, learning discriminative patterns from labeled normal and anomalous sequences.
Inspired by TextCNN, Lu et al.~\cite{lu2018detecting} apply multiple convolutional filters with different kernel sizes in parallel to capture local patterns at different granularities, achieving strong performance in supervised log anomaly detection.
LogRobust~\cite{zhang2019robust} combines semantic log-template representations with an attention-based recurrent model to improve robustness to unstable log events. 
LightLog~\cite{wang2022lightlog} improves efficiency by applying PCA-based dimensionality reduction to log semantic representations before using a lightweight temporal convolutional network to model sequential dependencies.
NeuralLog~\cite{le2021log} uses pretrained language models to encode the semantics of log events and adopts a Transformer-based encoder to capture contextual dependencies within log sequences, thereby improving anomaly classification.

Recent studies have also begun to explore the potential of large language models in log anomaly detection.
LogLLM~\cite{guan2024logllm} follows a semantic-enhanced framework similar to NeuralLog: it first uses pretrained language models such as BERT to encode log sequences, then employs a projector to align the extracted log representations with the input space of LLaMA, and finally applies parameter-efficient fine-tuning techniques such as LoRA for anomaly classification. 
Another line of work adapts GPT-2~\cite{lim2025adapting} to log anomaly detection through a simpler preprocessing pipeline, directly feeding the processed log-text sequences into GPT-2.





\subsection{Model Calibration}
Model calibration aims to make predictive confidence better reflect empirical correctness. Existing methods can be broadly grouped into three categories: regularization-based training, uncertainty estimation, and post-hoc calibration~\cite{gawlikowski2023survey}.

Regularization-based methods improve calibration by reducing overconfidence during training. Typical examples include label smoothing~\cite{muller2019does}, confidence penalty~\cite{pereyra2017regularizing}, mixup~\cite{thulasidasan2019mixup}, focal loss~\cite{mukhoti2020calibrating}, and class-balanced loss~\cite{cui2019class}. These methods can improve calibration, but require detector retraining and may alter the original detection behavior.

Uncertainty estimation methods quantify predictive uncertainty through approaches such as Monte Carlo dropout~\cite{gal2016dropout}, deep ensembles~\cite{lakshminarayanan2017simple}, and Bayesian neural networks~\cite{wilson2020bayesian}. However, they often require extra computation or multiple forward passes, making them less suitable for lightweight post-hoc calibration.

Post-hoc calibration methods adjust the output probabilities of a trained model without modifying its parameters. Representative methods include Temperature Scaling and Logistic Scaling~\cite{guo2017calibration}, Beta Scaling~\cite{kull2017beta}, and Selective Scaling~\cite{wang2023calibrating}. These methods improve confidence estimation through probability-level mappings or auxiliary calibration models, and serve as strong post-hoc calibration baselines.








\section{Preliminaries}

\subsection{Problem Formulation}
\label{sec:pro_formulation}

We study supervised log anomaly detection. The problem is formulated as a binary confidence calibration problem. Given an input log instance $x_i$ with binary indicator label $y_i \in \{0,1\}$, where $y_i=0$ denotes a normal sample and $y_i=1$ denotes an anomalous sample, a detector $f_\theta$ produces a scalar anomaly confidence score:

\begin{equation}
p_i = \sigma(f_\theta(x_i)), \quad p_i \in (0,1),
\end{equation}

where $\sigma(\cdot)$ denotes the sigmoid function and $p_i$ represents the predicted confidence that $x_i$ is anomalous. The final prediction is obtained by applying a decision threshold $\tau$:

\begin{equation}
\hat{y}_i =
\begin{cases}
1, & \text{if } p_i \geq \tau,\\
0, & \text{if } p_i < \tau.
\end{cases}
\end{equation}

Based on the relationship between the predicted label $\hat{y}_i$ and the ground-truth label $y_i$, predictions are partitioned into four categories: true negatives (TN), false positives (FP), true positives (TP), and false negatives (FN). This formulation enables systematic analysis of prediction confidence and calibration behavior across correct and incorrect anomaly predictions.

\subsection{Calibration Metrics}

Besides accuracy and F1 score, we evaluate model reliability using several widely adopted confidence calibration metrics for log anomaly detection. For example, negative log-likelihood (NLL) evaluates the quality of probabilistic predictions by measuring the likelihood assigned to the ground-truth labels. Lower NLL indicates better calibrated confidence estimates. To provide a comprehensive assessment of calibration quality, we further consider the following metrics that capture confidence--accuracy alignment, probabilistic prediction quality, and confidence behavior on both correct and incorrect predictions.

\textbf{Expected Calibration Error (ECE).}
ECE~\cite{naeini2015obtaining} measures the discrepancy between
prediction confidence and empirical accuracy by partitioning predictions
into $M$ fixed-width confidence bins:
\begin{equation}
    \mathrm{ECE} = \sum_{m=1}^{M} \frac{|B_m|}{N}
    \left| \mathrm{acc}(B_m) - \mathrm{conf}(B_m) \right|,
\end{equation}
where $B_m$ denotes the $m$-th confidence bin, $N$ is the total number
of samples, and $\mathrm{acc}(B_m)$ and $\mathrm{conf}(B_m)$ represent
the empirical accuracy and average confidence within the bin,
respectively. 

\textbf{Brier Score (BS).}
BS measures the mean squared error between predicted
probabilities and ground-truth labels:
\begin{equation}
    \mathrm{BS} = \frac{1}{N} \sum_{i=1}^{N} (p_i - y_i)^2.
\end{equation}
Unlike ECE, which evaluates calibration by comparing confidence with empirical accuracy, BS measures overall probability quality by directly penalizing probability estimation errors. As a result, BS reflects both calibration and class discrimination. However, similar to ECE, BS aggregates over all predictions and may therefore overlook confidence behavior on rare but critical mispredictions.

\textbf{Confidence on Error (CoE).}
CoE measures the average confidence assigned to
misclassified samples:
\begin{equation}
    \mathrm{CoE} = \frac{1}{|\mathcal{E}|} \sum_{i \in \mathcal{E}} c_i,
\end{equation}
where $\mathcal{E} = \{i \mid \hat{y}_i \neq y_i\}$ denotes the set of
incorrectly classified samples and $c_i$ is the confidence score of
sample $i$. A high CoE indicates that the model is systematically
overconfident on its mistakes --- a particularly concerning failure mode
in anomaly detection, where missed abnormal cases carry high operational
cost. Unlike ECE and BS, which aggregate over all predictions,
CoE isolates confidence behavior specifically on failure cases, providing
a targeted diagnostic that the other metrics do not directly expose. 
Conversely, Confidence on Correct (CoC) measures the average confidence assigned to correctly classified samples.

\section{Empirical Analysis of Confidence Calibration in Log Anomaly Detection}








\subsection{History Length}

We analyze the influence of sliding-window history size while fixing the stride to 1. As shown in Figure~\ref{fig:ablation_studies}(a), increasing the history length produces only modest variations in predictive performance, with Acc remaining above 98\% and F1 remaining above 0.84 across all settings. Although ECE and Brier score exhibit noticeable changes as the context length varies, CoE remains consistently high, ranging from 0.90 to 0.98. Interestingly, larger history windows reduce CoE more substantially than they improve conventional calibration metrics. While ECE increases from 0.13 to 0.17 as the history length grows from 5 to 30, CoE decreases from 0.98 to 0.90. This discrepancy suggests that average calibration metrics do not adequately reflect confidence behavior on erroneous predictions. Even when additional contextual information improves reliability, the model continues to assign excessively high confidence to incorrect predictions. Therefore, longer temporal context alone is insufficient to resolve the overconfidence problem.

\subsection{LoRA Rank}

Figure~\ref{fig:ablation_studies}(b) presents the impact of LoRA rank on calibration behavior. Increasing the LoRA rank substantially improves predictive performance, with F1 increasing from 0.61 at rank 4 to 0.94 at rank 32. Accuracy exhibits a similar trend, indicating that larger adaptation capacity enables the model to capture richer task-specific patterns. However, calibration-related metrics remain largely unchanged. ECE and Brier score stay nearly constant across all LoRA ranks, while CoE remains extremely high, approaching 1.0 regardless of adaptation capacity. These results suggest that increasing model expressiveness primarily improves predictive accuracy but provides limited benefit for mitigating overconfident errors. Even highly accurate models continue to assign near-maximal confidence to misclassified samples. This observation further demonstrates that improving predictive performance alone is insufficient for reliable confidence estimation.

\subsection{Model Complexity}

Figure~\ref{fig:ablation_studies}(c) investigates the effect of model complexity by varying the number of trainable parameters in a multilayer perceptron. Performance metrics remain nearly identical across all model sizes, with Acc and F1 consistently exceeding 0.999. Likewise, ECE and Brier score remain extremely small, suggesting excellent calibration according to conventional metrics. Surprisingly, CoE increases monotonically as model size grows, rising from 0.90 to nearly 0.99. This finding indicates that larger models become increasingly confident on their mistakes despite exhibiting near-perfect conventional calibration scores. The result highlights a potential trade-off between model capacity and confidence reliability, where additional representational power may amplify overconfidence on rare erroneous predictions. Consequently, model scaling alone does not improve confidence quality and may even exacerbate error confidence.

\subsection{Solutions for Imbalances}

Figure~\ref{fig:ablation_studies}(d) compares several commonly used strategies for addressing class imbalance. Traditional sampling-based methods, including over-sampling, under-sampling, and balanced sampling, produce only modest improvements in calibration behavior. Although these approaches slightly reduce CoE relative to the baseline, error confidence remains consistently high. Among all methods, class-weighted loss achieves the largest reduction in CoE, decreasing error confidence from 0.996 to 0.718 while simultaneously reducing ECE and Brier score. Focal loss also improves calibration relative to the baseline but remains less effective than class weighting. These results suggest that imbalance-aware optimization is more effective than data-level resampling for reducing overconfidence. Nevertheless, even the strongest baseline continues to exhibit substantial error confidence, indicating that class imbalance treatment alone cannot fully address the overconfidence problem.

\subsection{Class-wise Calibration}

Figure~\ref{fig:supp_bgl_classwise} compares calibration behavior between normal and abnormal logs. A consistent class-wise disparity is observed across all detectors. In terms of calibration metrics, abnormal logs exhibit substantially larger ECE, NLL, and Brier scores than normal logs, indicating that confidence estimates for anomaly predictions are considerably less reliable. The gap is particularly pronounced for LightLog, NeuralLog, and GPT-2, where abnormal-class calibration errors exceed those of the normal class by more than an order of magnitude. The confidence decomposition in Figure~\ref{fig:supp_bgl_classwise}(d) further reveals the source of this disparity. While both normal and abnormal correctly classified samples (Nor. CoC and Abn. CoC) generally receive high confidence, incorrectly classified abnormal samples (Abn. CoE) also receive surprisingly high confidence scores, often exceeding 0.9. In contrast, confidence on incorrectly classified normal samples (Nor. CoE) remains substantially lower and closer to the uncertainty region. This indicates that language model-based log anomaly detectors tend to be particularly overconfident when making anomaly-related mistakes, especially false negatives. These observations suggest that calibration errors are not uniformly distributed across classes. Instead, the primary calibration challenge arises from overconfident mispredictions involving anomalous logs. Consequently, improving confidence reliability for anomaly predictions is likely to yield the greatest gains in overall calibration quality, motivating the anomaly-focused route-wise calibration strategy adopted by LoRD.

\subsection{Characterizing Error Confidence}

The preceding analyses reveal a consistent pattern across different architectures, hyperparameters, and training strategies. Although ECE and Brier score often indicate moderate calibration performance, confidence on erroneous predictions (CoE) remains persistently high, frequently exceeding 0.9. This suggests that models can appear well calibrated on average while still assigning near-maximal confidence to incorrect predictions. This discrepancy arises because conventional calibration metrics are dominated by correctly classified samples and therefore provide limited insight into confidence behavior on rare prediction errors. For log anomaly detection, such overconfident failures are particularly problematic because they may lead to missed anomalies or incorrect operational decisions. These observations consistently show that \textbf{CoE remains a persistent challenge} regardless of model architecture, history window length, adaptation rank, model complexity, or imbalance-handling strategy. This finding suggests that confidence on erroneous predictions is an important yet largely overlooked aspect of reliability in log anomaly detection. Consequently, we introduce CoE as a complementary reliability metric and develop a reliability-aware calibration framework that explicitly optimizes confidence quality on misclassified samples.

\begin{figure*}[t]
    \centering
    \includegraphics[width=0.85\textwidth]{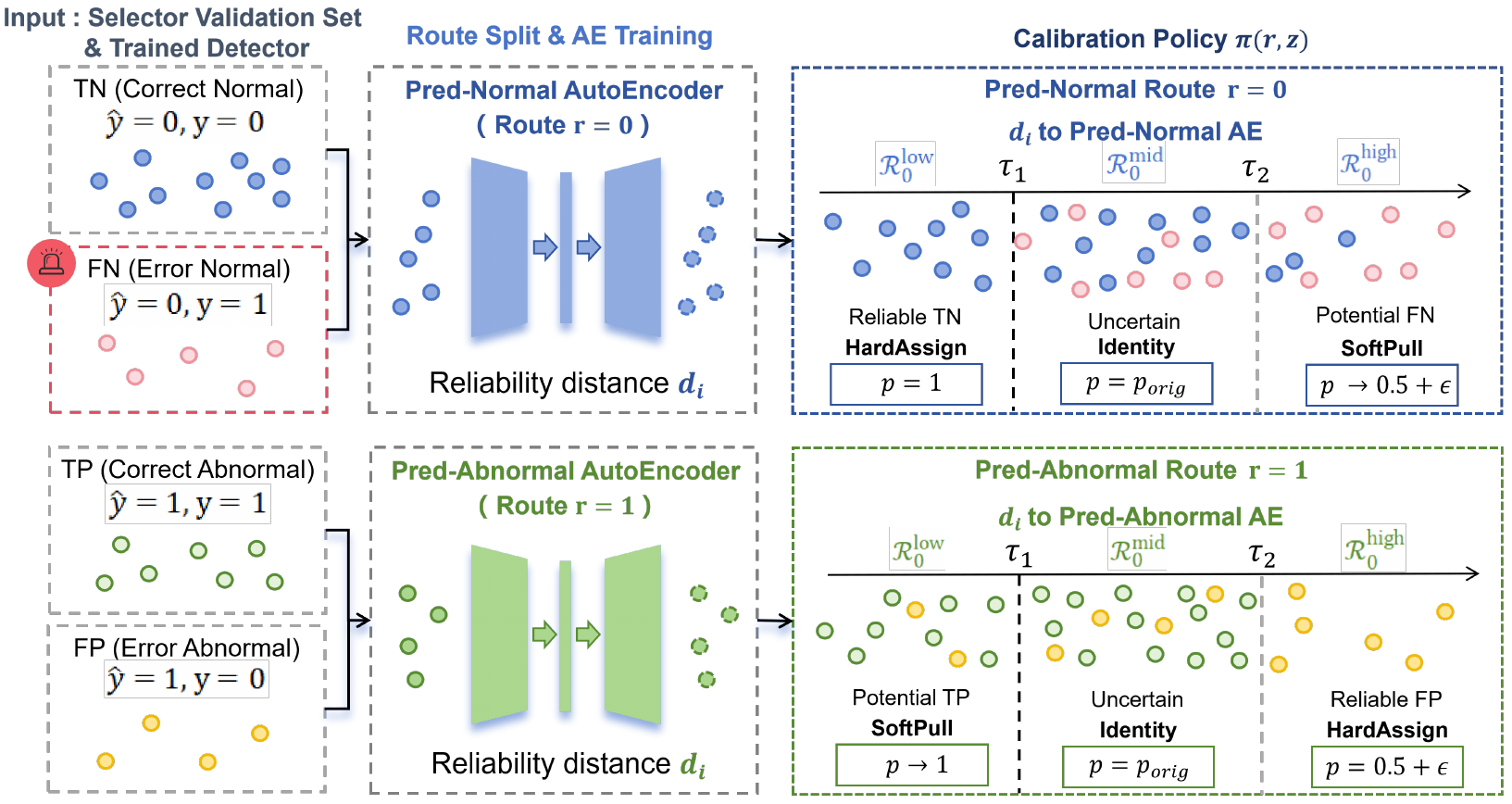}
    \caption{Overview of LoRD with AE. Correctly classified validation samples are separated by prediction route to train route-specific AE. Reliability distances are then used to partition predictions into route-aware regions, which are subsequently mapped to confidence calibration policies.}
    \label{fig:algorithm}
\end{figure*}

\section{LoRD : Reliability-Aware Calibration}








In this section, we present \textbf{Log Reconstruction and Distance (LoRD)}, a lightweight calibration framework designed to improve reliability in log anomaly detection. Building on the observation that confidence on erroneous predictions remains persistently high across different detectors and training settings, LoRD estimates prediction reliability from latent representations and uses the resulting reliability scores to recalibrate confidence. Specifically, LoRD combines route-wise reconstruction modeling, asymmetric threshold and margin selection, and selective confidence adjustment to better align prediction confidence with underlying prediction reliability. Unlike global post-hoc calibration, LoRD uses reconstruction distance as a route-specific reliability signal and performs asymmetric confidence adjustment according to the predicted route and reliability region, explicitly targeting overconfident errors while preserving confidence on reliable predictions.

Given an input log sequence $x_i$, a trained anomaly detector produces a hidden representation $\mathbf{h}_i \in \mathbb{R}^{d}$, a predicted probability vector $\mathbf{p}_i$, and a predicted label $\hat{y}_i \in \{0,1\}$. The true label is denoted as $y_i$, where $0$ indicates normal and $1$ indicates anomalous. Our goal is to recalibrate the detector confidence so that high-risk predictions, especially anomalous samples incorrectly predicted as normal, receive lower confidence while preserving the detector's original prediction labels.



\subsection{Representation Learning}

LoRD estimates prediction reliability through route-specific conditional representation learning in the detector latent space. Let $\mathbf{h}_i$ denote the hidden representation of sample $i$, and let $\hat{y}_i,y_i\in{0,1}$ denote the predicted and ground-truth labels, respectively. We posit that correctly classified samples exhibit route-dependent latent distributions, i.e.,
\begin{equation}
p(\mathbf{h}\mid \hat{y}=0,y=0)
\neq
p(\mathbf{h}\mid \hat{y}=1,y=1).
\end{equation}
Consequently, fitting a single reliability model to samples from both routes may obscure route-specific reliability structures, making prediction errors more difficult to distinguish from reliable predictions. To address this issue, LoRD learns route-specific reliability models that capture the conditional latent distributions of reliable predictions:
\begin{equation}
\mathcal{M}_r
\sim
p(\mathbf{h}\mid \hat{y}=r, y=r),
\qquad
r\in\{0,1\},
\end{equation}
where $\mathcal{M}_r$ denotes the reliable latent manifold associated with prediction route $r$. For each route, LoRD learns a representation function $f_r(\cdot)$ and a reconstruction function $g_r(\cdot)$. Given a sample assigned to route $\hat{y}_i$, the corresponding route-specific model generates a reconstructed latent representation:
\begin{equation}
\hat{\mathbf{h}}_i
=
g_{\hat{y}_i}
\!\left(
f_{\hat{y}_i}(\mathbf{h}_i)
\right).
\end{equation}

In this work, $f_r(\cdot)$ and $g_r(\cdot)$ are implemented as encoder--decoder networks, although alternative representation learning architectures can be incorporated. We estimate the reliability distance of a sample using its reconstruction error:
\begin{equation}
d_i
=
\left\|
\mathbf{h}_i-\hat{\mathbf{h}}_i
\right\|_2^{2},
\end{equation}
where $d_i$ measures the deviation of $\mathbf{h}_i$ from the route-specific reliable manifold $\mathcal{M}_{\hat{y}_i}$.
A small reliability distance indicates that the sample is well aligned with the reliable conditional distribution of its prediction route, whereas a large distance suggests atypical latent behavior. Consequently, false negatives tend to deviate from the reliable normal manifold, while false positives tend to deviate from the reliable anomaly manifold. The resulting reliability distance provides a unified measure of prediction reliability and serves as the basis for subsequent confidence calibration.

\subsection{Route-Aware Reliability Partitioning}

LoRD constructs route-specific calibration regions based on the reliability distance from the corresponding reliable manifold. To capture route-specific error patterns, LoRD learns two thresholds, $\tau_r^{(1)}<\tau_r^{(2)}$, for each prediction route and partitions samples into low-, mid-, and high-distance regions:
\begin{equation}
\begin{aligned}
\mathcal{R}_r^{\mathrm{low}}
&=
\{d_i \le \tau_r^{(1)}\},\\
\mathcal{R}_r^{\mathrm{mid}}
&=
\{\tau_r^{(1)} < d_i \le \tau_r^{(2)}\},\\
\mathcal{R}_r^{\mathrm{high}}
&=
\{d_i > \tau_r^{(2)}\}.
\end{aligned}
\end{equation}

Here, $\tau_r^{(1)}$ and $\tau_r^{(2)}$ are selected on the validation set according to the route-specific target recall $R_r$ and flagged rate $\rho_r$. In practice, $R_r$ and $\rho_r$ are selected according to the operational risk tolerance of the application. Larger $R_r$ increases the likelihood of identifying unreliable predictions, whereas smaller $\rho_r$ limits unnecessary calibration of reliable samples. The normal route $(r=0)$ prioritizes false-negative coverage, while the abnormal route $(r=1)$ prioritizes preserving reliable anomaly alarms.

This partition reflects the route-dependent reliability structure. 
In the normal route, samples with small reliability distance are more likely to be reliable true negatives, whereas samples with large distance are more likely to be false negatives. 
In the abnormal route, samples with small reliability distance are more likely to be reliable true positives, whereas samples with large distance are more likely to be false positives. 
Samples in the mid-distance region are treated as uncertain and are left for the reject option in the subsequent calibration policy.



\subsection{Calibration Policies}

Based on the route-aware calibration map, LoRD assigns a calibration action to each route-region combination, as summarized in Table~\ref{tab:policy_lookup}. Predictions in the uncertain region are left unchanged through a reject option, while predictions in the reliable and high-risk regions undergo confidence adjustment according to their estimated log risk. The policy is designed to increase confidence for samples that are more likely to be safe logs and suppress confidence for samples that are more likely to be anomalous logs.



Based on Table~\ref{tab:policy_lookup}, LoRD defines a route-region calibration policy
$\pi(r,z)$, where $r$ denotes the prediction route and $z$ denotes the reliability-distance region. For notational simplicity, hereafter, we use $p_i$ to denote the confidence assigned to the detector's original prediction, rather than the anomaly-class probability defined in Section \ref{sec:pro_formulation}.
LoRD is label-preserving: it recalibrates the confidence score but does not recompute the predicted label after calibration.

The Identity policy leaves $p_i$ unchanged. 
For the remaining regions, LoRD either directly assigns a target confidence through HardAssign or gradually moves $p_i$ toward a target confidence through SoftPull. 
The target confidence $q_{r,z}$ is determined by the selected route-region policy. 
Specifically, the normal low-distance region uses $q_{r,z}=1$, the normal high-distance region uses $q_{r,z}\rightarrow0.5+\epsilon$, the abnormal low-distance region uses $q_{r,z}\rightarrow1$, and the abnormal high-distance region uses $q_{r,z}=0.5+\epsilon$.

For SoftPull, the adjustment strength is determined by the distance to the nearest calibration boundary:
\begin{equation}
\alpha_i
=
1-\exp\!\left(
-\frac{\Delta_i}{s}
\right),
\qquad
\alpha_i\in[0,1],
\end{equation}
where $\Delta_i$ denotes the boundary distance and $s$ is a scale parameter determined by the corresponding margin. 
The recalibrated confidence is computed as
\begin{equation}
p_i^{\mathrm{new}}
=
(1-\alpha_i)p_i
+
\alpha_i q_{r,z},
\end{equation}
where $q_{r,z}$ is specified by the selected calibration policy. 
Thus, samples close to the boundary are only mildly adjusted, while samples farther away are pulled more strongly toward the target confidence. 
For HardAssign, LoRD directly sets $p_i^{\mathrm{new}}=q_{r,z}$.



\begin{table}[t]
\centering
\caption{Route-region calibration policy map $\pi$.}
\label{tab:policy_lookup}

\footnotesize
\renewcommand{\arraystretch}{1.12}
\setlength{\tabcolsep}{3.5pt}

\resizebox{\columnwidth}{!}{
\begin{tabular}{lccc}
\toprule
\textbf{Route} 
& $\mathcal{R}_r^{\mathrm{low}}$ 
& $\mathcal{R}_r^{\mathrm{mid}}$ 
& $\mathcal{R}_r^{\mathrm{high}}$ \\
\midrule
Normal $(r=0)$
& HardAssign: $p=1$
& Identity
& SoftPull: $p\rightarrow 0.5+\epsilon$ \\

Abnormal $(r=1)$
& SoftPull: $p\rightarrow 1$
& Identity
& HardAssign: $p=0.5+\epsilon$ \\
\bottomrule
\end{tabular}
}
\end{table}

\subsection{Algorithm}

We summarize the calibration framework in Algorithm~\ref{alg:lord}. LoRD performs route-specific confidence calibration by modeling the latent representations of reliable predictions separately for the normal and abnormal routes. First, the base detector is applied to a validation set to obtain predicted labels, confidence scores, and hidden representations. For each route $r\in\{0,1\}$, correctly classified samples are treated as route-reliable instances, while misclassified samples form the route-error set. A route-specific autoencoder is then trained on the latent representations of reliable samples to learn the characteristic feature distribution of trustworthy predictions. Reconstruction distances are computed for both reliable and error samples, and these distances are used to construct low-, medium-, and high-reliability regions according to predefined route-specific targets. During inference, each test sample is assigned to its predicted route, and its reconstruction distance is evaluated using the corresponding route autoencoder. The resulting distance serves as a reliability indicator that is mapped through a calibration policy to adjust the detector's original confidence score, producing calibrated confidence scores that better reflect prediction reliability. In deployment, the calibrated confidence can serve as a reliability signal for downstream decision making; for example, low-confidence normal predictions can be flagged for further inspection rather than being treated as reliable normal events.

\begin{algorithm}[t]
\caption{LoRD Calibration}
\label{alg:lord}
\small
\begin{algorithmic}[1]
\Require Detector $f_\theta$, validation set $\mathcal{D}_{sel}$, test set $\mathcal{D}_{test}$, route targets $\{(R_r,\rho_r)\}_{r\in\{0,1\}}$, calibration policy $\pi$
\Ensure Calibrated confidence scores $\mathbf{p}_i^{new}$

\State Obtain $(\hat{y}_i,y_i,\mathbf{p}_i,\mathbf{h}_i)$ from $f_\theta(\mathcal{D}_{sel})$
\State $\hat{y}_i,y_i\in\{0,1\}$, where $0$ is normal and $1$ is abnormal

\For{$r\in\{0,1\}$}
    \State Define route-reliable set $\mathcal{T}_r=\{i:\hat{y}_i=r,\;y_i=r\}$
    \State Define route-error set $\mathcal{E}_r=\{i:\hat{y}_i=r,\;y_i\neq r\}$
    \State Train route autoencoder $(f_r,g_r)$ on $\{\mathbf{h}_i:i\in\mathcal{T}_r\}$
    \State Compute $d_i^r=\|\mathbf{h}_i-g_r(f_r(\mathbf{h}_i))\|_2^2$ for $i\in\mathcal{T}_r\cup\mathcal{E}_r$
    \State Update $\mathcal{R}_r^{low}$, $\mathcal{R}_r^{mid}$, $\mathcal{R}_r^{high}$ under $(R_r,\rho_r)$ with $\mathcal{E}_r$
\EndFor

\For{$i\in\mathcal{D}_{test}$}
    \State Obtain $(\hat{y}_i,\mathbf{p}_i,\mathbf{h}_i)$ from $f_\theta(x_i)$
    \State Set route $r\leftarrow\hat{y}_i$
    \State Compute $d_i^r=\|\mathbf{h}_i-g_r(f_r(\mathbf{h}_i))\|_2^2$

    \State Calibrate $\mathbf{p}_i$ with $\pi$ given $d_i^r$
\EndFor

\end{algorithmic}
\end{algorithm}

\section{Experiments}

\begin{table*}[h]
\centering
\caption{Calibration performance measured by CoE on anomalous log detection across benchmark datasets and representative detectors. Reported values denote mean CoE with standard deviation over multiple runs, where lower values indicate better performance. The best and second-best results in each row are highlighted in bold and underlined italics.}
\label{tab:overall_calibration}
\footnotesize
\renewcommand{\arraystretch}{0.88}
\setlength{\tabcolsep}{3.2pt}

\begin{tabular*}{\textwidth}{@{\extracolsep{\fill}}llccccccc@{}}
\toprule
\textbf{Dataset} & \textbf{Model}
& \textbf{Uncal} & \textbf{TempS} & \textbf{LogS} & \textbf{BetaS}
& \textbf{SeleS} & \textbf{Ens.} & \textbf{LoRD} \\
\midrule

\multirow{5}{*}{BGL}
& TextCNN
& \ms{0.977}{0.003} & \ms{0.976}{0.004} & \ms{0.988}{0.002} & \ms{0.989}{0.003}
& \ms{0.968}{0.005} & \second{\ms{0.966}{0.005}} & \lord{0.540}{0.006} \\
& LogRobust
& \ms{0.987}{0.009} & \ms{0.973}{0.016} & \ms{0.988}{0.004} & \ms{0.988}{0.003}
& \ms{0.907}{0.098} & \second{\ms{0.748}{0.163}} & \lord{0.577}{0.092} \\
& LightLog
& \second{\ms{0.829}{0.022}} & \ms{0.983}{0.007} & \ms{0.967}{0.009} & \ms{0.979}{0.004}
& \ms{0.831}{0.015} & \ms{0.998}{0.000} & \lord{0.702}{0.036} \\
& NeuralLog
& \ms{0.990}{0.002} & \ms{0.975}{0.007} & \ms{0.974}{0.007} & \ms{0.988}{0.001}
& \second{\ms{0.847}{0.016}} & \ms{0.985}{0.001} & \lord{0.509}{0.002} \\
& GPT-2
& \ms{0.998}{0.000} & \ms{0.992}{0.000} & \ms{0.993}{0.000} & \second{\ms{0.991}{0.000}}
& \ms{1.000}{0.000} & \ms{0.998}{0.000} & \lord{0.566}{0.002} \\

\midrule

\multirow{5}{*}{Spirit}
& TextCNN
& \ms{0.890}{0.055} & \ms{0.862}{0.051} & \second{\ms{0.776}{0.091}} & \ms{0.833}{0.056}
& \ms{0.858}{0.039} & \ms{0.926}{0.009} & \lord{0.556}{0.022} \\
& LogRobust
& \ms{0.957}{0.007} & \ms{0.936}{0.005} & \second{\ms{0.929}{0.009}} & \ms{0.950}{0.011}
& \ms{0.930}{0.021} & \ms{0.951}{0.010} & \lord{0.506}{0.004} \\
& LightLog
& \ms{0.959}{0.014} & \ms{0.983}{0.007} & \ms{0.967}{0.009} & \ms{0.979}{0.004}
& \second{\ms{0.929}{0.010}} & \ms{0.948}{0.029} & \lord{0.514}{0.014} \\
& NeuralLog
& \ms{0.996}{0.003} & \ms{0.975}{0.007} & \ms{0.974}{0.007} & \ms{0.988}{0.001}
& \second{\ms{0.972}{0.009}} & \ms{0.991}{0.006} & \lord{0.508}{0.007} \\
& GPT-2
& \ms{0.936}{0.000} & \ms{1.000}{0.000} & \ms{0.982}{0.009} & \ms{0.942}{0.000}
& \ms{0.998}{0.001} & \second{\ms{0.936}{0.001}} & \lord{0.500}{0.000} \\

\midrule

\multirow{5}{*}{Liberty}
& TextCNN
& \ms{0.979}{0.002} & \ms{0.972}{0.003} & \second{\ms{0.881}{0.025}} & \ms{0.933}{0.013}
& \ms{0.949}{0.016} & \ms{0.967}{0.002} & \lord{0.506}{0.002} \\
& LogRobust
& \ms{0.937}{0.050} & \second{\ms{0.930}{0.052}} & \ms{0.944}{0.020} & \ms{0.979}{0.007}
& \ms{0.958}{0.036} & \ms{0.952}{0.038} & \lord{0.502}{0.002} \\
& LightLog
& \ms{0.959}{0.015} & \ms{0.960}{0.020} & \ms{0.925}{0.044} & \ms{0.979}{0.007}
& \second{\ms{0.861}{0.022}} & \ms{0.916}{0.030} & \lord{0.505}{0.001} \\
& NeuralLog
& \ms{0.995}{0.005} & \ms{0.994}{0.006} & \second{\ms{0.982}{0.007}} & \ms{0.995}{0.003}
& \ms{0.995}{0.004} & \ms{0.995}{0.006} & \lord{0.683}{0.018} \\
& GPT-2
& \ms{1.000}{0.000} & \ms{0.993}{0.003} & \ms{0.994}{0.000} & \second{\ms{0.988}{0.001}}
& \ms{1.000}{0.000} & \ms{1.000}{0.000} & \lord{0.590}{0.004} \\

\midrule

\multirow{5}{*}{Thunderbird}
& TextCNN
& \ms{0.852}{0.020} & \ms{0.884}{0.007} & \ms{0.910}{0.027} & \ms{0.999}{0.000}
& \second{\ms{0.708}{0.052}} & \ms{0.737}{0.013} & \lord{0.561}{0.040} \\
& LogRobust
& \ms{0.926}{0.033} & \ms{0.929}{0.048} & \ms{0.902}{0.093} & \ms{0.995}{0.004}
& \ms{0.898}{0.036} & \second{\ms{0.706}{0.076}} & \lord{0.619}{0.123} \\
& LightLog
& \second{\ms{0.656}{0.016}} & \ms{0.973}{0.036} & \ms{0.975}{0.038} & \ms{0.999}{0.001}
& \ms{0.905}{0.063} & \ms{0.723}{0.048} & \lord{0.500}{0.000} \\
& NeuralLog
& \second{\ms{0.852}{0.087}} & \ms{0.947}{0.061} & \ms{0.951}{0.062} & \ms{0.991}{0.005}
& \ms{0.902}{0.087} & \ms{0.881}{0.060} & \lord{0.516}{0.021} \\
& GPT-2
& \ms{1.000}{0.000} & \ms{1.000}{0.000} & \ms{1.000}{0.000} & \second{\ms{0.994}{0.010}}
& \ms{1.000}{0.000} & \ms{1.000}{0.000} & \lord{0.534}{0.008} \\

\bottomrule
\end{tabular*}
\end{table*}

\subsection{Datasets}

We evaluate our method on four large-scale supercomputing log datasets: \textbf{BGL}~\cite{oliner2007supercomputers}, \textbf{Spirit}~\cite{oliner2007supercomputers}, \textbf{Liberty}~\cite{oliner2007supercomputers}, and \textbf{Thunderbird}~\cite{oliner2007supercomputers}, covering anomaly ratios ranging from 0.49\% to 32.01\%. Following prior work, we use 4.7M log messages from BGL, the first 5M log messages from Spirit and Liberty, and 10M log messages from Thunderbird. To prevent information leakage, each dataset is chronologically partitioned into non-overlapping training, detector-validation, selector-validation, and test splits with a ratio of $7:0.5:0.5:2$. The training split is used to optimize the base detector, the detector-validation split is used for early stopping and checkpoint selection, and the selector-validation split is used for calibrator training and threshold selection. When route-specific validation errors are unavailable in the selector-validation split, the detector-validation split is used as a fallback. The test split is reserved exclusively for final evaluation.

\newcommand{\best}[1]{\begingroup\bfseries\boldmath #1\endgroup}

\begin{table*}[h]
\centering
\caption{Ablation study of LoRD on BGL, Spirit, and Liberty with LogRobust and NeuralLog as base detectors. 
Abn. CoC and Abn. CoE denote the average confidence on correctly and incorrectly classified abnormal samples, respectively.}
\label{tab:ablation_lord_route_abnormal}

\footnotesize
\renewcommand{\arraystretch}{0.92}
\setlength{\tabcolsep}{3.2pt}

\begin{tabular*}{\textwidth}{@{\extracolsep{\fill}}llcccccc@{}}
\toprule
\textbf{Detector} 
& \textbf{Architecture}
& \multicolumn{2}{c}{\textbf{BGL}}
& \multicolumn{2}{c}{\textbf{Spirit}}
& \multicolumn{2}{c}{\textbf{Liberty}} \\
\cmidrule(lr){3-4} \cmidrule(lr){5-6} \cmidrule(lr){7-8}
& 
& \textbf{Abn. CoC}$\uparrow$
& \textbf{Abn. CoE}$\downarrow$
& \textbf{Abn. CoC}$\uparrow$
& \textbf{Abn. CoE}$\downarrow$
& \textbf{Abn. CoC}$\uparrow$
& \textbf{Abn. CoE}$\downarrow$ \\
\midrule

\multirow{3}{*}{LogRobust}
& SingleAE
& \ms{0.562}{0.066} & \ms{0.582}{0.038}
& \ms{0.556}{0.041} & \ms{0.516}{0.011}
& \ms{0.740}{0.049} & \ms{0.506}{0.002} \\

& Route-0-Only
& \best{\ms{0.993}{0.005}} & \ms{0.542}{0.002}
& \ms{0.999}{0.001} & \ms{0.583}{0.018}
& \best{\ms{0.999}{0.001}} & \ms{0.617}{0.050} \\

& \textbf{LoRD}
& \ms{0.990}{0.003}
& \best{\ms{0.541}{0.002}}
& \best{\ms{0.999}{0.000}}
& \best{\ms{0.508}{0.003}}
& \ms{0.986}{0.008}
& \best{\ms{0.502}{0.001}} \\

\midrule

\multirow{3}{*}{NeuralLog}
& SingleAE
& \ms{0.549}{0.052} & \ms{0.610}{0.023}
& \ms{0.647}{0.111} & \ms{0.543}{0.059}
& \ms{0.783}{0.033} & \ms{0.788}{0.200} \\

& Route-0-Only
& \ms{0.998}{0.001} & \best{\ms{0.506}{0.002}}
& \ms{0.998}{0.001} & \ms{0.575}{0.059}
& \best{\ms{1.000}{0.000}} & \ms{0.716}{0.044} \\

& \textbf{LoRD}
& \best{\ms{0.998}{0.001}}
& \best{\ms{0.506}{0.002}}
& \best{\ms{0.999}{0.001}}
& \best{\ms{0.511}{0.008}}
& \best{\ms{1.000}{0.000}}
& \best{\ms{0.680}{0.047}} \\

\bottomrule
\end{tabular*}
\vspace{-0.1in}
\end{table*}



\subsection{Log Anomaly Detectors}
We evaluate five representative supervised log anomaly detectors, including TextCNN~\cite{lu2018detecting}, LogRobust~\cite{zhang2019robust}, LightLog~\cite{wang2022lightlog}, NeuralLog~\cite{le2021log}, and GPT-2~\cite{lim2025adapting}, covering conventional deep learning and language model-based approaches. Unless otherwise specified, all detectors are evaluated at the log-sequence level using a sliding-window configuration with a history length of 10 and stride of 1.

\subsection{Baseline Calibration Methods}

\begin{table*}[h]
\centering
\caption{Ablation study of LoRD on Spirit and Liberty with LogRobust as the base detector. CoC and CoE denote the average confidence on correctly and incorrectly classified samples, respectively. Lower $D$ and higher $C$ indicate better overall calibration quality.}
\label{tab:ablation_lord}
\footnotesize
\renewcommand{\arraystretch}{0.90}
\setlength{\tabcolsep}{3.0pt}

\begin{tabular*}{\textwidth}{@{\extracolsep{\fill}}lcccccccccccc@{}}
\toprule
& \multicolumn{6}{c}{\textbf{Spirit}}
& \multicolumn{6}{c}{\textbf{Liberty}} \\
\cmidrule(lr){2-7}
\cmidrule(lr){8-13}
\textbf{Method}
& Nor. CoC$\uparrow$
& Nor. CoE$\downarrow$
& Abn. CoC$\uparrow$
& Abn. CoE$\downarrow$
& $D\downarrow$
& $C\uparrow$
& Nor. CoC$\uparrow$
& Nor. CoE$\downarrow$
& Abn. CoC$\uparrow$
& Abn. CoE$\downarrow$
& $D\downarrow$
& $C\uparrow$ \\
\midrule

w/o Reject
& 0.899 & 0.957 & \textbf{1.000} & 0.504 & 0.468 & 0.754
& 0.856 & 0.922 & \textbf{0.994} & \textbf{0.501} & 0.446 & 0.759 \\

w/o Soft
& 0.908 & 0.905 & \textbf{1.000} & \textbf{0.503} & 0.415 & 0.776
& 0.879 & 0.843 & 0.988 & \textbf{0.501} & 0.364 & 0.796 \\

\textbf{LoRD}
& \textbf{0.942} & \textbf{0.901} & \textbf{1.000} & 0.509
& \textbf{0.405} & \textbf{0.783}
& \textbf{0.916} & \textbf{0.812} & 0.986 & 0.502
& \textbf{0.323} & \textbf{0.817} \\

\bottomrule
\end{tabular*}
\end{table*}

\begin{table}
\centering
\caption{Component ablation of LoRD using NeuralLog as the base detector.
CoC and CoE denote the confidence on correctly and incorrectly classified
abnormal samples, respectively. ``w/o Reject'' and ``w/o Soft'' remove the
reject region and SoftPull calibration, respectively. Lower $D$ and higher
$C$ indicate better overall confidence reliability.}
\label{tab:ablation_lord_component}

\footnotesize
\setlength{\tabcolsep}{3pt}
\renewcommand{\arraystretch}{0.9}

\begin{tabular*}{\columnwidth}{@{\extracolsep{\fill}}lcccccccc@{}}
\toprule
&
\multicolumn{4}{c}{\textbf{Spirit}}
&
\multicolumn{4}{c}{\textbf{Liberty}}
\\
\cmidrule(lr){2-5}
\cmidrule(lr){6-9}

\textbf{Method}
& CoC$\uparrow$
& CoE$\downarrow$
& $D\downarrow$
& $C\uparrow$
& CoC$\uparrow$
& CoE$\downarrow$
& $D\downarrow$
& $C\uparrow$ \\
\midrule

w/o Reject
& \textbf{1.000} & \textbf{0.507} & 0.508 & 0.736
& \textbf{1.000} & \textbf{0.572} & 0.450 & 0.757 \\

w/o Soft
& \textbf{1.000} & \textbf{0.507} & 0.506 & 0.741
& \textbf{1.000} & 0.602 & 0.395 & 0.790 \\

\textbf{LoRD}
& \textbf{1.000} & 0.513 & \textbf{0.495} & \textbf{0.747}
& \textbf{1.000} & 0.680 & \textbf{0.354} & \textbf{0.814} \\

\bottomrule
\end{tabular*}
\vspace{-0.1in}
\end{table}

We compare LoRD against five representative post-hoc calibration methods. \textbf{Temperature Scaling (TempS)}~\cite{guo2017calibration} calibrates confidence using $\mathbf{p}_i^{\mathrm{TS}}=\mathrm{softmax}(\mathbf{z}_i/T)$, where $T$ is optimized on a validation set. \textbf{Logistic Scaling (LogS)}~\cite{guo2017calibration} generalizes TS through a learnable affine transformation, $\mathbf{p}_i^{\mathrm{LS}}=\mathrm{softmax}(\mathbf{w}\odot\mathbf{z}_i+\mathbf{b})$. \textbf{Beta Scaling (BetaS)}~\cite{kull2017beta} performs probability-level calibration using $p_i^{\mathrm{BS}}=\sigma(a\log p_i+b\log(1-p_i)+c)$. \textbf{Selective Scaling (SeleS)}~\cite{wang2023calibrating} applies separate temperatures to selected high-risk and low-risk predictions based on output-level signals such as confidence, entropy, or logit margin. \textbf{Ensembling (Ens.)}~\cite{wang2023calibrating} averages predictions from multiple independently trained detectors, $\mathbf{p}_i^{\mathrm{Ens}}=\frac{1}{K}\sum_{k=1}^{K}\mathbf{p}_i^{(k)}$. All calibration parameters are optimized using the validation set.

\subsection{Results}

\subsubsection{Performance}
The main experiments show that LoRD consistently improves calibration reliability for anomalous log detection across datasets and base detectors. In Table~\ref{tab:overall_calibration}, LoRD obtains the lowest CoE in nearly every dataset–model combination, indicating that it effectively suppresses overconfident misclassifications. Compared to standard post-hoc methods such as Temperature Scaling, Logistic Scaling, Beta Scaling, Selective Scaling, and Ensembling, LoRD yields substantially lower error confidence while preserving reliable anomaly alarms. This behavior is pronounced for language-model-based and high-capacity detectors, where LoRD reduces CoE from values near one to values close to 0.5 in many settings.

LoRD also maintains high confidence on correctly classified anomalous samples, as reflected by its robustness in CoC values. This demonstrates that the proposed route-wise reconstruction mechanism selectively lowers confidence on risky predictions while retaining trustworthy anomaly detections. These results confirm that LoRD provides a task-driven calibration trade-off that is well suited for supervised log anomaly detection.

\subsubsection{Ablation study}

Table~\ref{tab:ablation_lord_route_abnormal} first evaluates the necessity of route-specific reliability modeling in LoRD. Replacing the route-specific autoencoders with a single shared autoencoder (SingleAE) consistently degrades abnormal calibration across all datasets and detectors. For example, with LogRobust on Spirit, Abn. CoE increases from \(0.508\) to \(0.516\), while with NeuralLog on Liberty it increases from \(0.680\) to \(0.788\). These results suggest that predicted-normal and predicted-abnormal samples follow distinct reliability patterns in the latent space. A shared reconstruction model must simultaneously characterize both distributions, which may blur route-dependent reliability structures and reduce its ability to distinguish reliable predictions from potential errors. In contrast, route-specific modeling allows each autoencoder to learn a more homogeneous reliability manifold, resulting in improved separation between reliable and unreliable predictions. This observation is consistent with our hypothesis that normal and abnormal prediction routes exhibit different latent reliability characteristics and therefore benefit from separate modeling.

We next evaluate the contributions of the two remaining components of LoRD: the reject region mechanism and the distance aware soft calibration strategy. 
To jointly characterize calibration across correct and erroneous predictions, we additionally define $D$ and $C$ as summary measures of confidence–correctness alignment.
Let $\mathbf{x}=(x_1,\ldots,x_m)$ denote a set of calibration objectives. For each objective $x_j$, we define a target value $t_j$, where $t_j=1$ for correct prediction confidence metrics and $t_j=0.5$ for error confidence metrics. The objective deviation vector is defined as
$\mathbf{z}
=
\left(
|x_1-t_1|,
\ldots,
|x_m-t_m|
\right).$ We then define the aggregate calibration distance as
$D=\|\mathbf{z}\|_2$. To obtain a normalized relative score, we further define
$C=
\frac{\|\mathbf{z}-\mathbf{z}^{-}\|_2}
{\|\mathbf{z}\|_2+\|\mathbf{z}-\mathbf{z}^{-}\|_2}$
,where $\mathbf{z}^{-}$ denotes the worst achievable deviation vector. Smaller $D$ and larger $C$ indicate better overall calibration quality. 


Tables~\ref{tab:ablation_lord} and~\ref{tab:ablation_lord_component}  evaluate the reject region and distance-aware SoftPull calibration using LogRobust and NeuralLog, respectively. Across both detectors and datasets, the complete LoRD achieves the best overall confidence trade-off, consistently yielding the lowest $D$ and highest $C$. For example, on Spirit with LogRobust, LoRD achieves $D=0.405$ and $C=0.783$, outperforming w/o Reject ($D=0.468$, $C=0.754$) and w/o Soft ($D=0.415$, $C=0.776$). Although individual abnormal-class confidence metrics may vary, the consistent improvements in $D$ and $C$ demonstrate that the two components provide complementary benefits in balancing confidence preservation and error suppression.

\section{Conclusion}

We propose LoRD, a lightweight post-hoc calibration framework for supervised log anomaly detection. By training separate autoencoders for predicted-normal and predicted-anomalous routes, LoRD captures route-specific reliability signals from detector latent representations and selectively adjusts high-risk predictions while preserving original labels. Experiments on multiple benchmark datasets and representative detectors demonstrate that LoRD reduces confidence on misclassified anomalous samples, preserves trustworthy anomaly alarms, and outperforms existing calibration baselines. These results highlight the effectiveness of task-oriented, route-aware calibration for reliable log anomaly detection.

\section{Limitations}

LoRD has several limitations. First, its threshold and margin selection depend on route-specific validation errors, so the calibration boundary may be less stable when false negatives or false positives are scarce. Second, as a task-oriented calibration method, LoRD prioritizes confidence--correctness alignment for high-risk errors rather than uniformly optimizing aggregate calibration metrics, which may result in trade-offs in ECE, NLL, or Brier score. Third, LoRD requires access to detector hidden representations and introduces an additional autoencoder forward pass during calibration, limiting its applicability to black-box systems and introducing additional inference overhead. Future work will investigate robust calibration under sparse validation errors, stronger distribution shifts, multi-class diagnosis, and online deployment efficiency.

\section{Acknowledgment}
This research was supported by the National Natural Science Foundation of China (No.62376023).

\bibliographystyle{IEEEtran}
\bibliography{custom}

@inproceedings{han2022skdlog,
  title={Skdlog: self-knowledge distillation-based cnn for abnormal log detection},
  author={Han, Ningning and Lu, Siyang and Wang, Dongdong and Wang, Mingquan and Tan, Xiaoman and Wei, Xiang},
  booktitle={2022 IEEE Smartworld, Ubiquitous Intelligence \& Computing, Scalable Computing \& Communications, Digital Twin, Privacy Computing, Metaverse, Autonomous \& Trusted Vehicles (SmartWorld/UIC/ScalCom/DigitalTwin/PriComp/Meta)},
  pages={796--805},
  year={2022},
  organization={IEEE}
}

@article{lu2023ssdlog,
  title={SSDLog: a semi-supervised dual branch model for log anomaly detection},
  author={Lu, Siyang and Han, Ningning and Wang, Mingquan and Wei, Xiang and Lin, Zaichao and Wang, Dongdong},
  journal={World Wide Web},
  volume={26},
  number={5},
  pages={3137--3153},
  year={2023},
  publisher={Springer}
}

@article{lu2023black,
  title={Black-box attacks against log anomaly detection with adversarial examples},
  author={Lu, Siyang and Wang, Mingquan and Wang, Dongdong and Wei, Xiang and Xiao, Sizhe and Wang, Zhiwei and Han, Ningning and Wang, Liqiang},
  journal={Information Sciences},
  volume={619},
  pages={249--262},
  year={2023},
  publisher={Elsevier}
}

@inproceedings{nedelkoski2020self,
  title={Self-attentive classification-based anomaly detection in unstructured logs},
  author={Nedelkoski, Sasho and Bogatinovski, Jasmin and Acker, Alexander and Cardoso, Jorge and Kao, Odej},
  booktitle={2020 IEEE international conference on data mining (ICDM)},
  pages={1196--1201},
  year={2020},
  organization={IEEE}
}

@inproceedings{le2022log,
  title={Log-based anomaly detection with deep learning: How far are we?},
  author={Le, Van-Hoang and Zhang, Hongyu},
  booktitle={Proceedings of the 44th international conference on software engineering},
  pages={1356--1367},
  year={2022}
}

@article{landauer2024critical,
  title={A critical review of common log data sets used for evaluation of sequence-based anomaly detection techniques},
  author={Landauer, Max and Skopik, Florian and Wurzenberger, Markus},
  journal={Proceedings of the ACM on Software Engineering},
  volume={1},
  number={FSE},
  pages={1354--1375},
  year={2024},
  publisher={ACM New York, NY, USA}
}

@inproceedings{zhu2023loghub,
  title={Loghub: A large collection of system log datasets for ai-driven log analytics},
  author={Zhu, Jieming and He, Shilin and He, Pinjia and Liu, Jinyang and Lyu, Michael R},
  booktitle={2023 IEEE 34th International Symposium on Software Reliability Engineering (ISSRE)},
  pages={355--366},
  year={2023},
  organization={IEEE}
}

@inproceedings{wu2023towards,
  title={Towards reliable rare category analysis on graphs via individual calibration},
  author={Wu, Longfeng and Lei, Bowen and Xu, Dongkuan and Zhou, Dawei},
  booktitle={Proceedings of the 29th ACM SIGKDD Conference on Knowledge Discovery and Data Mining},
  pages={2629--2638},
  year={2023}
}

@inproceedings{du2017deeplog,
  title={Deeplog: Anomaly detection and diagnosis from system logs through deep learning},
  author={Du, Min and Li, Feifei and Zheng, Guineng and Srikumar, Vivek},
  booktitle={Proceedings of the 2017 ACM SIGSAC conference on computer and communications security},
  pages={1285--1298},
  year={2017}
}

@inproceedings{meng2019loganomaly,
  title={Loganomaly: Unsupervised detection of sequential and quantitative anomalies in unstructured logs.},
  author={Meng, Weibin and Liu, Ying and Zhu, Yichen and Zhang, Shenglin and Pei, Dan and Liu, Yuqing and Chen, Yihao and Zhang, Ruizhi and Tao, Shimin and Sun, Pei and others},
  booktitle={Ijcai},
  volume={19},
  number={7},
  pages={4739--4745},
  year={2019}
}

@inproceedings{yang2021plelog,
  title={Plelog: Semi-supervised log-based anomaly detection via probabilistic label estimation},
  author={Yang, Lin and Chen, Junjie and Wang, Zan and Wang, Weijing and Jiang, Jiajun and Dong, Xuyuan and Zhang, Wenbin},
  booktitle={2021 IEEE/ACM 43rd International Conference on Software Engineering: Companion Proceedings (ICSE-Companion)},
  pages={230--231},
  year={2021},
  organization={IEEE}
}

@inproceedings{lu2018detecting,
  title={Detecting anomaly in big data system logs using convolutional neural network},
  author={Lu, Siyang and Wei, Xiang and Li, Yandong and Wang, Liqiang},
  booktitle={2018 IEEE 16th Intl Conf on Dependable, Autonomic and Secure Computing, 16th Intl Conf on Pervasive Intelligence and Computing, 4th Intl Conf on Big Data Intelligence and Computing and Cyber Science and Technology Congress (DASC/PiCom/DataCom/CyberSciTech)},
  pages={151--158},
  year={2018}
}

@inproceedings{zhang2019robust,
  title={Robust log-based anomaly detection on unstable log data},
  author={Zhang, Xu and Xu, Yong and Lin, Qingwei and Qiao, Bo and Zhang, Hongyu and Dang, Yingnong and Xie, Chunyu and Yang, Xinsheng and Cheng, Qian and Li, Ze and others},
  booktitle={Proceedings of the 2019 27th ACM joint meeting on European software engineering conference and symposium on the foundations of software engineering},
  pages={807--817},
  year={2019}
}

@article{wang2022lightlog,
  title={LightLog: A lightweight temporal convolutional network for log anomaly detection on the edge},
  author={Wang, Zumin and Tian, Jiyu and Fang, Hui and Chen, Liming and Qin, Jing},
  journal={Computer Networks},
  volume={203},
  pages={108616},
  year={2022},
  publisher={Elsevier}
}

@inproceedings{le2021log,
  title={Log-based anomaly detection without log parsing},
  author={Le, Van-Hoang and Zhang, Hongyu},
  booktitle={2021 36th IEEE/ACM International Conference on Automated Software Engineering (ASE)},
  pages={492--504},
  year={2021},
  organization={IEEE}
}

@article{guan2024logllm,
  title={Logllm: Log-based anomaly detection using large language models},
  author={Guan, Wei and Cao, Jian and Qian, Shiyou and Gao, Jianqi and Ouyang, Chun},
  journal={arXiv preprint arXiv:2411.08561},
  year={2024}
}

@inproceedings{lim2025adapting,
  title={Adapting large language models for parameter-efficient log anomaly detection},
  author={Lim, Ying Fu and Zhu, Jiawen and Pang, Guansong},
  booktitle={Pacific-Asia Conference on Knowledge Discovery and Data Mining},
  pages={325--337},
  year={2025},
  organization={Springer}
}

@article{gawlikowski2023survey,
  title={A survey of uncertainty in deep neural networks},
  author={Gawlikowski, Jakob and Tassi, Cedrique Rovile Njieutcheu and Ali, Mohsin and Lee, Jongseok and Humt, Matthias and Feng, Jianxiang and Kruspe, Anna and Triebel, Rudolph and Jung, Peter and Roscher, Ribana and others},
  journal={Artificial intelligence review},
  volume={56},
  number={Suppl 1},
  pages={1513--1589},
  year={2023},
  publisher={Springer}
}

@article{muller2019does,
  title={When does label smoothing help?},
  author={M{\"u}ller, Rafael and Kornblith, Simon and Hinton, Geoffrey E},
  journal={Advances in neural information processing systems},
  volume={32},
  year={2019}
}

@article{pereyra2017regularizing,
  title={Regularizing neural networks by penalizing confident output distributions},
  author={Pereyra, Gabriel and Tucker, George and Chorowski, Jan and Kaiser, {\L}ukasz and Hinton, Geoffrey},
  journal={arXiv preprint arXiv:1701.06548},
  year={2017}
}

@article{thulasidasan2019mixup,
  title={On mixup training: Improved calibration and predictive uncertainty for deep neural networks},
  author={Thulasidasan, Sunil and Chennupati, Gopinath and Bilmes, Jeff A and Bhattacharya, Tanmoy and Michalak, Sarah},
  journal={Advances in neural information processing systems},
  volume={32},
  year={2019}
}

@article{mukhoti2020calibrating,
  title={Calibrating deep neural networks using focal loss},
  author={Mukhoti, Jishnu and Kulharia, Viveka and Sanyal, Amartya and Golodetz, Stuart and Torr, Philip and Dokania, Puneet},
  journal={Advances in neural information processing systems},
  volume={33},
  pages={15288--15299},
  year={2020}
}

@inproceedings{cui2019class,
  title={Class-balanced loss based on effective number of samples},
  author={Cui, Yin and Jia, Menglin and Lin, Tsung-Yi and Song, Yang and Belongie, Serge},
  booktitle={Proceedings of the IEEE/CVF conference on computer vision and pattern recognition},
  pages={9268--9277},
  year={2019}
}

@inproceedings{gal2016dropout,
  title={Dropout as a bayesian approximation: Representing model uncertainty in deep learning},
  author={Gal, Yarin and Ghahramani, Zoubin},
  booktitle={international conference on machine learning},
  pages={1050--1059},
  year={2016},
  organization={PMLR}
}

@article{lakshminarayanan2017simple,
  title={Simple and scalable predictive uncertainty estimation using deep ensembles},
  author={Lakshminarayanan, Balaji and Pritzel, Alexander and Blundell, Charles},
  journal={Advances in neural information processing systems},
  volume={30},
  year={2017}
}

@article{wilson2020bayesian,
  title={Bayesian deep learning and a probabilistic perspective of generalization},
  author={Wilson, Andrew G and Izmailov, Pavel},
  journal={Advances in neural information processing systems},
  volume={33},
  pages={4697--4708},
  year={2020}
}

@inproceedings{guo2017calibration,
  title={On calibration of modern neural networks},
  author={Guo, Chuan and Pleiss, Geoff and Sun, Yu and Weinberger, Kilian Q},
  booktitle={International conference on machine learning},
  pages={1321--1330},
  year={2017},
  organization={PMLR}
}

@inproceedings{kull2017beta,
  title={Beta calibration: a well-founded and easily implemented improvement on logistic calibration for binary classifiers},
  author={Kull, Meelis and Silva Filho, Telmo and Flach, Peter},
  booktitle={Artificial intelligence and statistics},
  pages={623--631},
  year={2017},
  organization={PMLR}
}

@inproceedings{wang2023calibrating,
  title={On calibrating semantic segmentation models: Analyses and an algorithm},
  author={Wang, Dongdong and Gong, Boqing and Wang, Liqiang},
  booktitle={Proceedings of the IEEE/CVF Conference on Computer Vision and Pattern Recognition},
  pages={23652--23662},
  year={2023}
}

@inproceedings{naeini2015obtaining,
  title={Obtaining well calibrated probabilities using bayesian binning},
  author={Naeini, Mahdi Pakdaman and Cooper, Gregory and Hauskrecht, Milos},
  booktitle={Proceedings of the AAAI conference on artificial intelligence},
  volume={29},
  number={1},
  year={2015}
}

@inproceedings{oliner2007supercomputers,
  title={What supercomputers say: A study of five system logs},
  author={Oliner, Adam and Stearley, Jon},
  booktitle={37th annual IEEE/IFIP international conference on dependable systems and networks (DSN'07)},
  pages={575--584},
  year={2007},
  organization={IEEE}
}

\clearpage

\end{document}